\pdfoutput=1
\documentclass[11pt,a4paper]{article}
\usepackage[hyperref]{emnlp-ijcnlp-2019}
\usepackage{times}
\usepackage{amsmath}
\usepackage{amsfonts}
\usepackage{amssymb}
\usepackage{latexsym}
\usepackage{booktabs}
\usepackage{color} 
\usepackage{setspace}
\usepackage{latexsym}
\usepackage{multirow}
\usepackage{url}
\usepackage{graphicx} 
\usepackage{subfigure}
\aclfinalcopy 

\title{Concurrent Parsing of Constituency and Dependency}

\author{Junru Zhou , Shuailiang Zhang,  Hai Zhao\thanks{$\ $  Corresponding author. This paper was partially supported by National
Key Research and Development Program of China (No.
2017YFB0304100) and key projects of Natural Science Foundation of
China (No. U1836222 and No. 61733011)} \\
  Department of Computer Science and Engineering \\
  Key Lab of Shanghai Education Commission \\
  for Intelligent Interaction and Cognitive Engineering \\
  MoE Key Lab of Artificial Intelligence, AI Institute \\
  Shanghai Jiao Tong University, Shanghai, China \\
  {\tt \{zhoujunru,zsl123\}@sjtu.edu.cn, zhaohai@cs.sjtu.edu.cn} 
  }

\date{}

\begin{document}
\maketitle
\begin{abstract}
    Constituent and dependency representation for syntactic structure share a lot of linguistic and computational characteristics, this paper thus makes the first attempt by introducing a new model that is capable of parsing constituent and dependency at the same time, so that lets either of the parsers enhance each other. Especially, we evaluate the effect of different shared network components and empirically verify that dependency parsing may be much more beneficial from constituent parsing structure.
    The proposed parser achieves new state-of-the-art performance for both parsing tasks, constituent and dependency on PTB and CTB benchmarks.
\end{abstract}

\section{Introduction}

\begin{figure}[t!]
    \centering
    \subfigure[Constituent structure]{
        \label{Fig.sub.1}
        \includegraphics[width=1.3in]{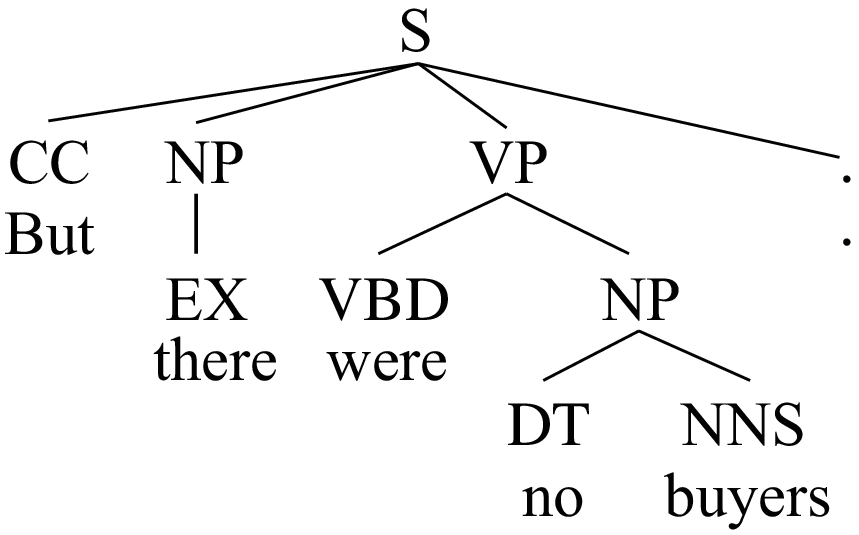}
    }
    \subfigure[Dependency structure]{
        \label{Fig.sub.2}
        \includegraphics[width=1.5in]{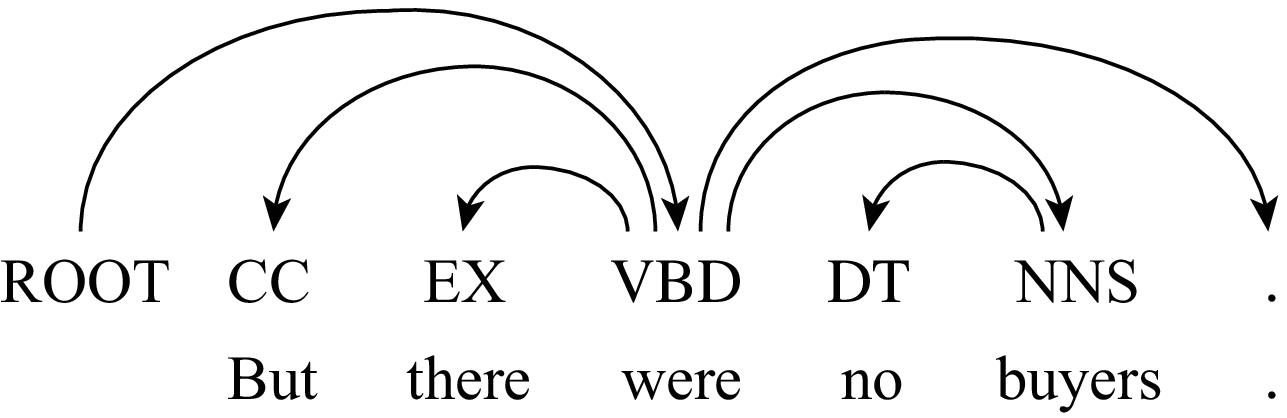}
    }
    \caption{Constituent and dependency structures.}
    \label{fig1}
\end{figure}

Constituent and dependency are two typical syntactic structure representation forms as shown in Figure \ref{fig1}, which have been well studied from both linguistic and computational perspective  \cite{chomsky1981lectures, Leeman2001Bresnan}. In earlier time, linguists and NLP researchers discussed how to encode lexical dependencies in phrase structures, like Tree-adjoining grammar (TAG) \cite{joshi1997tree} and head-driven phrase structure grammar (HPSG) \cite{pollard1994head}.

Typical dependency treebanks are usually converted from constituent treebanks, though they may be independently annotated as well for the same languages. Meanwhile, constituent parsing can be accurately converted to dependencies (SD) representation by grammatical rules or machine learning methods \cite{Marieffe06generatingtyped,MaW10-4146}. 
Such mutual convertibility shows a close relation between constituent and dependency representation for the same sentence.
Thus, it is a natural idea to study the relationship
between constituent and dependency structures, and the joint learning of constituent and dependency parsing \cite{CollinsP97-1003,CharniakA00-2018,KleinP04,CharniakP05-1022,FarkasW11-2924,GreenW12-0503,RenCombine2013,XuP14-1021,YoshikawaP17-1026,strzyz-etal-2019-sequence}.

\begin{figure*}[t!]
    \centering
    \includegraphics[width=5in]{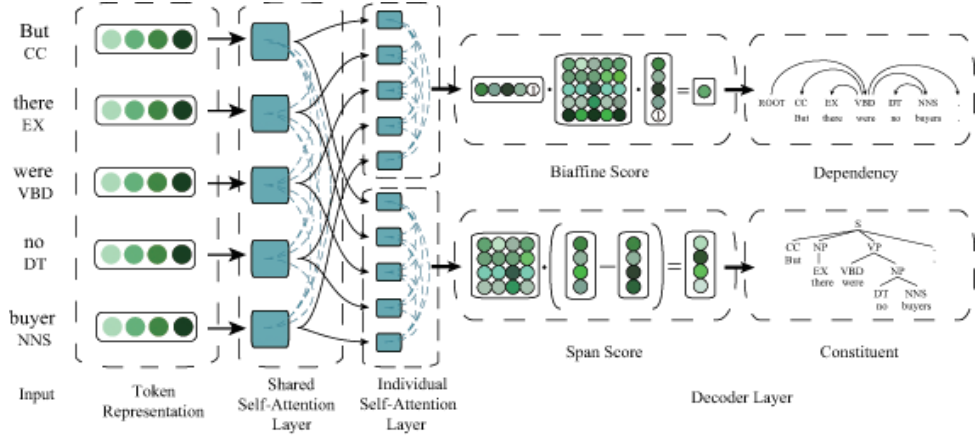}
    
    \caption{The framework of our joint learning model.}
    \label{fig2}
\end{figure*}

For further exploit both strengths of the two representation forms for even better parsing, in this work, we propose a new model that is capable of synchronously parsing constituent and dependency.

Multitask learning (MTL) is a natural solution in neural models for multiple inputs and multiple outputs, which is adopted in this work to decode constituent and dependency in a single model. \cite{SogaardP16} indicates that when tasks are sufficiently similar, especially with syntactic nature, MTL would be useful. In contrast to previous work on deep MTL \cite{Collobert2011Natural, Hashimoto2016A}, our model focuses on more related tasks and benefits from the strong inherent relation. 
At last, our model is evaluated on two benchmark treebanks for both constituent and dependency parsing. The empirical results show that our parser reaches new state-of-the-art for all parsing tasks.

\section{Our Model}

Using an encoder-decoder backbone, our model may be regarded as an extension of the constituent parsing model of \cite{Kitaev-2018-SelfAttentive} as shown in Figure \ref{fig2}. The difference is that in our model both constituent and dependency parsing share the same token representation and shared self-attention layers and each has its own individual Self-Attention Layers and subsequent processing layers. Our model includes four modules: token representation, self-attention encoder, constituent and dependency parsing decoder.

\subsection{Token Representation}

In our model, token representation $x_i$ is composed by character, word and part-of-speech (POS) embeddings.
For character-level representation, we explore two types of encoders, CharCNNs \cite{MaP16, ChiuQ16} and CharLSTM \cite{Kitaev-2018-SelfAttentive}, as both types have been verified their effectiveness.
For word-level representation, we concatenate randomly initialized and pre-trained word embeddings.
We consider two ways to compose the final token representation, summing and concatenation,
$x_i$=$x_{char}$+$x_{word}$+$x_{POS}$, $x_i$=[$x_{char}$;$x_{word}$;$x_{POS}$]. 
\label{Token Representation}

\subsection{Self-Attention Encoder}

The encoder in our model is adapted from \cite{Vaswani17} to factor explicit content and position information in the self-attention process \cite{Kitaev-2018-SelfAttentive}. The input matrices $X = [x_1, x_2, \dots , x_n ]$ in which $x_i$ is concatenated with position embedding are transformed by a self-attention encoder. We factor the model between content and position information both in self-attention sub-layer and feed-forward network, whose setting details follow \cite{Kitaev-2018-SelfAttentive}. We also try different numbers of shared self-attention layers in section \ref{shares component}. 

\subsection{Constituent Parsing Decoder}

The score $s(T)$ of the constituent parsing tree $T$ is to sum every scores of span ($i$, $j$) with label $l$,

$
s(T) = \sum_{(i,j,\ell)\in T} s(i, j, \ell).
$

The goal of constituent parser is to find the tree with the highest score:
$
\hat{T} = \arg\max_T s(T).
$
We use CKY-style algorithm to obtain the tree $\hat{T}$ in $O(n^3)$ time complexity \cite{Cocke1970Programming, Younger1975Recognition, Kasami1965AN,SternP17,Gaddy}.

This structured prediction problem is handled with satisfying the margin constraint:

$
s(T^*) \ge s(T) + \Delta (T,T^*),
$

\noindent where $T^*$ denote correct parse tree and $\Delta$ is the Hamming loss on labeled spans with a slight modification during the dynamic programming search. The objective function is the hinge loss,
$$J_1(\theta) = \max ( 0,\max_T[s(T) + \Delta (T,T^*)]-s(T^*) ).$$

\subsection{Dependency Parsing Decoder}

Similar to the constituent case, dependency parsing is to search over all possible trees to find the globally highest scoring tree.
We follow \cite{Dozat2017Deep} and \cite{ZhangE17} to predict a distribution over the possible head for each word and find the globally highest scoring tree conditional on the distribution of each word only during testing.

We use the biaffine attention mechanism \cite{Dozat2017Deep} between each word and the candidates of the parent node:

$\alpha_{ij} = h_i^TWg_j + U^Th_i + V^T g_j + b,$

\noindent where $h_i$ and $g_i$ are calculated by a distinct one-layer perceptron network.

Dependency parser is to minimize the negative log likelihood of the golden tree $Y$, which is implemented as cross-entropy loss:

$
J_2(\theta) = - \left(logP_{\theta}(h_i|x_i) +logP_{\theta}(l_i|x_i,h_i)\right),
$

\noindent where $P_{\theta}(h_i|x_i)$ is the probability of correct parent node $h_i$ for $x_i$, and $P_{\theta}(l_i|x_i,h_i)$ is the probability of the correct dependency label $l_i$ for the child-parent pair $(x_i,h_i)$.

During parsing, we use the first-order Eisner algorithm \cite{EisnerP96} to build projective trees.

\subsection{Joint training}

Our joint model synchronously predicts the dependency tree and the constituent tree over the same input sentence. The output of the self-attention encoder is sent to the different decoder to generate the different parse tree. Thus, the share components for two parsers include token representation layer and self-attention encoder. 

We jointly train the constituent and dependency parser for minimizing the overall loss:

$J_{model}(\theta) = J_1(\theta) + \lambda J_2(\theta),$

\noindent where $\lambda$ is a hyper-parameter to control the overall loss. The best performance can be achieved when $\lambda$ is set to 1.0, which turns out that both sides are equally important.

\section{Experiments}

We evaluate our model on two benchmark treebanks, English Penn Treebank (PTB) and Chinese Penn Treebank (CTB5.1) following standard data splitting \cite{ZhangD08,Liuandzhang2017B}. POS tags are predicted by the Stanford Tagger \cite{Toutanova:2003}.
For constituent parsing, we use the standard evalb{\footnote{http://nlp.cs.nyu.edu/evalb/}} tool to evaluate the F1 score. For dependency parsing, we apply Stanford basic dependencies (SD) representation \cite{Marieffe06generatingtyped} converted by the Stanford parser{\footnote{http://nlp.stanford.edu/software/lex-parser.html}}.
Following previous work \cite{Dozat2017Deep, Ma2018Stack}, we report the results without punctuations for both treebanks.

\subsection{Setup}

We use the same experimental settings as \cite{Kitaev-2018-SelfAttentive}.
For dependency parsing, we employ two 1024-dimensional multilayer perceptrons for learning specific representation and a 1024-dimensional parameter matrix for biaffine attention. 
We use 100D GloVe \cite{PenningtonD14-1162} for English and structured-skipgram \cite{LingN15-1142} embeddings for Chinese.

\begin{table}[t!]
    \begin{center}
    \small
    \resizebox{\linewidth}{!}{    
        \begin{tabular*}{\hsize}{@{}@{\extracolsep{\fill}}lccl@{}}
            \hline
            \bf Lexical Representation      &F1 &UAS &LAS\\
            \hline
            Word, POS, CharLSTM      &93.50 &95.41 &93.35 \\
            Word, POS, CharCNNs      &93.46 &95.43 &93.30 \\
            Word, CharLSTM   &\bf93.83 &\bf95.71 &\bf93.68 \\
            Word, CharCNNs           &93.70 &95.43 &93.35 \\
            POS, CharLSTM             &93.16 &95.06 &92.82 \\
            CharLSTM                        &93.80 &95.46 &93.46 \\
            \hline
        \end{tabular*}}
    \end{center}
    \caption{\label{table1} PTB dev set performance on representations.}
\end{table}

\begin{table}[t!]
    \begin{center}
        \small
        \resizebox{\linewidth}{!}{    
        \begin{tabular*}{\hsize}{@{}@{\extracolsep{\fill}}rlccl@{}}
        \hline
        \multicolumn{2}{c}{\bf Joint Component}  &F1 &UAS &LAS\\
        \hline
        \multicolumn{2}{c}{Separate}   &93.44 &94.59 &92.15\\
        \hline
        \multirow{5}{0.2cm}{Shared}&self-att 0layers   &93.76 &95.63 &93.56\\
        \multirow{5}{0.2cm}&self-att 2layers   &93.70 &95.48 &93.41\\
        \multirow{5}{0.2cm}&self-att 4layers   &93.59 &95.24 &93.12\\
        \multirow{5}{0.2cm}&self-att 6layers   &93.68 &95.50 &93.51\\
        \multirow{5}{0.2cm}&self-att 8layers   &\bf93.83 &\bf95.71 &\bf93.68\\
        \hline
        \end{tabular*}}
    \end{center}
    \caption{\label{table2} PTB dev set performance on joint component.}
\end{table}

\subsection{Ablation Studies}

All experiments in this subsection are running from token representation with summing setting.

\textbf{Token Representation} Different token representation combinations are evaluated in Table \ref{table1}. We find that CharLSTM performs a little better than CharCNNs. Moreover, POS tags on parsing performance show that predicted POS tags decreases parsing accuracy, especially without word information. 
If POS tags are replaced by word embeddings, the performance increases.
Finally, we apply word and CharLSTM as token representation setting for our full model{\footnote{We also evaluate POS tags on CTB which increases parsing accuracy, thus we employ the word, POS tags and CharLSTM as token representation setting for CTB.}}.

\begin{table}[t!]
    \begin{center}
    \resizebox{\linewidth}{!}{    
        \begin{tabular}{l|cc|cc}
        \hline
        \multirow{2}{*}{\bf Single Model} & \multicolumn{2}{c|}{English} & \multicolumn{2}{c}{Chinese} \\
        \cline{2-5}
        \multirow{2}{*}&UAS &LAS &UAS &LAS\\
        \hline
        \citet{Weiss2015Structured} &94.26 &92.41 &\_ &\_ \\
        \citet{Andor2016Globally} &94.61 &92.79 &\_ &\_\\
        \citet{Dozat2017Deep} &95.74 &94.08 &89.30 &88.23\\
        \citet{Ma2018Stack}  &95.87 &\textbf{94.19} &90.59  &89.29\\
        \hline
        Our model$^{\textit{Separate}}$(Sum)  &94.80 &92.45 &88.66 &86.58 \\
        Our model$^{\textit{Separate}}$(Concat)  &94.68 &92.32 &88.59 &86.17 \\
        Our model (Sum)   &95.90 &93.99 &\bf 90.89 & \bf 89.34\\
        Our model (Concat)   &\bf95.91 &93.86 &90.85  &89.28\\
        \hline
        \bf Pre-training \\
        \citet{WangD18-1311}(ELMo) &96.35 &\bf95.25 &\_ &\_ \\
        \hline
        Our model (ELMo)  & 96.82 & 94.91 &\_ &\_ \\
        Our model (BERT)  &\bf 96.88 & 95.12 &\_ &\_ \\
        \hline
        \hline
        \bf Ensemble \\
        \citet{ChoeD16-1257}  &95.9  &94.1 &\_ &\_      \\
        \citet{KuncoroE17-1117} &95.8  &94.6 &\_ &\_    \\
        \hline
        \end{tabular}}
    \end{center}
    \caption{\label{table3} Dependency parsing on PTB and CTB.}
\end{table}

\textbf{Shared Self-attention Layers} As our model providing two outputs from one input, there is a bifurcation setting for how much shared part should be determined. Both constituent and dependency parsers share token representation and 8 self-attention layers at most. Assuming that either parser always takes input information flow through 8 self-attention layers as shown in Figure \ref{fig2}, then the number of shared self-attention layers varying from 0 to 8 may reflect the shared degree in the model. When the number is set to 0, it indicates only token representation is shared for both parsers trained for the joint loss through each own 8 self-attention layers. When the number is set to less than 8, for example, 6, then it means that both parsers first shared 6 layers from token representation then have individual 2 self-attention layers.

For different numbers of shared layers, the results are in Table \ref{table2}. We respectively disable the constituent and the dependency parser to obtain a separate learning setting for both parsers in our model. The comparison in Table \ref{table2} indicates that even though without any shared self-attention layers, joint training of our model may significantly outperform separate learning mode. At last, the best performance is still obtained from sharing full 8 self-attention layers.

Besides, comparing UAS and LAS to F1 score, dependency parsing is shown more beneficial from our model which has more than 1\% gain in UAS and LAS from parsing constituent together.
\label{shares component}

\begin{table}[t!]
    \begin{center}
    \small
        \resizebox{\linewidth}{!}{
        \begin{tabular*}{\hsize}{@{}@{\extracolsep{\fill}}lccc@{}}
            \hline
            \bf Single Model               &LR &LP &F1\\
            \hline 
            \citet{SternP17}&93.2 &90.3 &91.8\\
            \citet{Gaddy} &91.76 &92.41 &92.08\\
            \citet{SternD17b} &92.57 &92.56 &92.56\\
            \citet{Kitaev-2018-SelfAttentive}  &93.20  &93.90 &93.55\\
            \hline
            Our model$^{\textit{Separate}}$ (Sum)  &92.92 &93.90 &93.41 \\
            Our model$^{\textit{Separate}}$ (Concat)  &93.26 &93.95 &93.60 \\
            Our model (Sum)   &93.52  &94.00 &93.76 \\
            Our model (Concat)    &\bf93.71  &\bf94.09 &\bf93.90 \\
            \hline
            \bf Pre-training \\
            \citet{Kitaev-2018-SelfAttentive}(ELMo) &94.85 &95.40 &95.13\\
            \citet{kitaev2018multilingual}(BERT) &95.46 &95.73 &95.59\\
            \hline
            Our model (ELMo)  &94.73 &95.25 &94.99\\
            Our model (BERT)  &\bf95.51 & \bf95.87 &\bf 95.69 \\
            \hline
            \hline
            \bf Ensemble \\
            \citet{LiuandZhang2017A} &\_ &\_ &94.2\\
            \citet{Fried2017Improving} &\_ &\_ &94.66\\
			\citet{kitaev2018multilingual} &95.51 &96.03 &95.77\\
            \hline
        \end{tabular*}}
    \end{center}
    \caption{\label{table4} Comparison of constituent parsing on PTB.}
\end{table}

\begin{table}[t!]
    \begin{center}
    \small
        \resizebox{\linewidth}{!}{
    \begin{tabular*}{\hsize}{@{}@{\extracolsep{\fill}}llccl@{}}
    \hline
    \bf Single Model              &LR &LP &F1\\
    \hline 
    \citet{Liuandzhang2017B} &85.9 &85.2 &85.5\\
    \citet{ShenP18}         &86.6 &86.4 &86.5 \\
    \citet{FriedP18}            &\_ &\_ &87.0 \\
	\citet{TengC18-1011}    &87.1 &87.5 &87.3\\
	\citet{kitaev2018multilingual} &91.55 &91.96 &91.75\\
    \hline 
    Our model$^{\textit{Separate}}$ (Sum) &91.35 &91.65 &91.50 \\
    Our model$^{\textit{Separate}}$ (Concat) &91.36 &92.02 &91.69\\
    Our model (Sum)    &\bf91.79  &\bf92.31 &\bf92.05 \\
    Our model (Concat)    &91.41  &92.03 &91.72 \\
    \hline
    \end{tabular*}}
    \end{center}
    \caption{\label{table5} Comparison of constituent parsing on CTB.}
\end{table}

\subsection{Main Results}

Tables \ref{table3}, \ref{table4} and \ref{table5} compare our model to existing state-of-the-art, in which indicator \textit{Separate} with our model shows the results of our model learning constituent or dependency parsing separately, (Sum) and (Concat) respectively represent the results with the indicated input token representation setting.
On PTB, our model achieves 93.90 F1 score of constituent parsing and 95.91 UAS and 93.86 LAS of dependency parsing.
On CTB, our model achieves a new state-of-the-art result on both constituent and dependency parsing.
The comparison again suggests that learning jointly in our model is superior to learning separately. 
In addition, we also augment our model with ELMo \cite{PetersN18-1202} or a larger version of BERT \cite{Jacobbert} as the sole token representation to compare with other pre-training models.
Since BERT is based on sub-word, we only take the last sub-word vector of the word in the last layer of BERT as our sole token representation $x_i$.
Moreover, our single model of BERT achieves competitive performance with other ensemble models.



\section{Conclusions}

This paper presents a joint model with the constituent and dependency parsing which achieves new state-of-the-art results on both Chinese and English benchmark treebanks.
Our ablation studies show that joint learning of both constituent and dependency is indeed superior to separate learning mode.
Also, experiments show that dependency parsing is much more beneficial from knowing the constituent structure. 
Our parser predicts phrase structure and head-word simultaneously which can be regarded as an effective HPSG \cite{pollard1994head} parser.
\bibliography{emnlp-ijcnlp-2019}
\bibliographystyle{acl_natbib}

\end{document}